\documentclass[letterpaper]{article} 
\usepackage{aaai2026}  
\usepackage{times}  
\usepackage{helvet}  
\usepackage{courier}  
\usepackage[hyphens]{url}  
\usepackage{graphicx} 
\usepackage{natbib}  
\usepackage{caption} 
\usepackage{algorithm}
\usepackage{algorithmic}

\usepackage{cite}
\usepackage{array}
\usepackage{amsmath, amssymb}
\usepackage{booktabs}
\usepackage{multirow}
\usepackage{placeins} 
\usepackage{makecell}
\usepackage{subcaption}
\usepackage{comment}

\usepackage{pifont}
\newcommand{\cmark}{\ding{51}} 
\newcommand{\xmark}{\ding{55}} 

\usepackage[table,xcdraw]{xcolor}

\newcolumntype{C}[1]{>{\centering\let\newline\\\arraybackslash\hspace{0pt}}m{#1}}
\newcolumntype{L}[1]{>{\raggedright\let\newline\\\arraybackslash\hspace{0pt}}m{#1}}

\usepackage{newfloat}
\usepackage{listings}
\DeclareCaptionStyle{ruled}{labelfont=normalfont,labelsep=colon,strut=off} 
\floatstyle{ruled}
\newfloat{listing}{tb}{lst}{}
\floatname{listing}{Listing}
\title{CLUENet: Cluster Attention Makes Neural Networks Have Eyes}
\author{
Xiangshuai Song\textsuperscript{\rm 1}\equalcontrib,
Jun-Jie Huang\textsuperscript{\rm 1}\equalcontrib,
Tianrui Liu\textsuperscript{\rm 1}\thanks{Corresponding author.},
Ke Liang\textsuperscript{\rm 1}\footnotemark[2],
Chang Tang\textsuperscript{\rm 2}
}
\affiliations{
\textsuperscript{\rm 1}College of Computer Science and Technology, National University of Defense Technology, Changsha, China\\
\textsuperscript{\rm 2}School of Software Engineering, Huazhong University of Science and Technology, Wuhan, China\\
sxs@nudt.edu.cn, jjhuang@nudt.edu.cn, trliu@nudt.edu.cn, liangke200694@126.com, tangchang@hust.edu.cn
}

\begin{document}

\maketitle

\begin{abstract}
Despite the success of convolution- and attention-based models in vision tasks, their rigid receptive fields and complex architectures limit their ability to model irregular spatial patterns and hinder interpretability, thereby posing challenges for tasks requiring high model transparency. Clustering paradigms offer promising interpretability and flexible semantic modeling, but suffer from limited accuracy, low efficiency, and gradient vanishing during training. To address these issues, we propose the CLUster attEntion Network (CLUENet), a transparent deep architecture for visual semantic understanding. Specifically, we introduce three key innovations, including (i) a Global and Soft Feature Aggregation with a Temperature-Scaled Cosine Attention for capturing long-range dependencies and a Gated Fusion Mechanism for enhanced local modeling, (ii) Hard and Shared Feature Dispatching, and (iii) an Improved Cluster Pooling Block. These enhancements significantly improve both classification performance and visual interpretability. Experiments on CIFAR-100 and Mini-ImageNet demonstrate that CLUENet outperforms existing clustering methods and mainstream visual models, offering a compelling balance of accuracy, efficiency, and transparency. 
\end{abstract}

\begin{links}
    \link{Code}{https://github.com/52KunKun/CLUENet}
\end{links}


\section{Introduction}
%
%
%
%
Deep Neural Networks have become the primary approach for image analysis, and various deep architectures have been proposed and differ in their modeling of intrinsic image structures. Early deep learning methods predominantly relied on the convolutional paradigm~{\cite{ref1,ref2,ref3,ref4,ref5,ref6,ref7}}. These Convolutional Neural Networks (CNNs) extract features by applying shared-weight kernels locally via sliding windows, inherently offering translation equivariance and local context modeling~{\cite{ref1}}. 
With the rise of Transformers, the attention-based paradigm~{\cite{ref8,ref9,ref10,ref11,ref12}} has emerged as a powerful alternative for image analysis. 
Transformers leverage global modeling capabilities to extract features through adaptive weight allocation. 
However, this comes at the cost of significant computational complexity.

\begin{figure}[!t]
	\centering
	\includegraphics[width=\columnwidth]{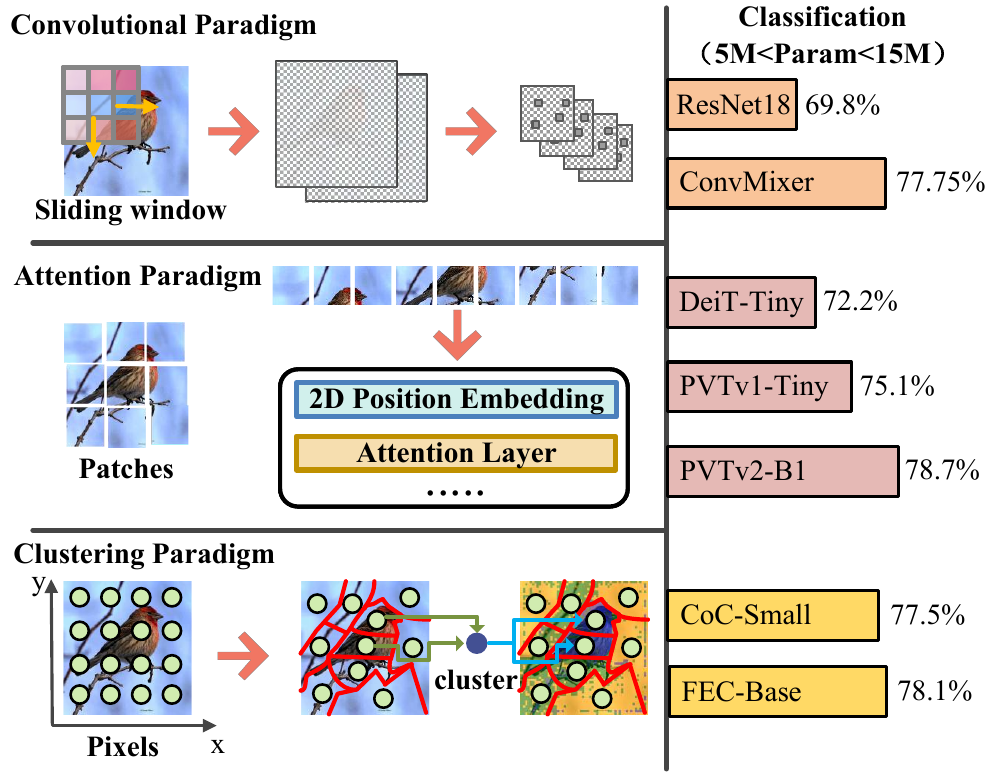}
	\caption{Different paradigms of visual models. Top: convolutional paradigms using convolution for local information modeling. Middle: attention paradigms leveraging positional embeddings and attention for global feature modeling. Bottom: clustering paradigms employing coordinate-guided clustering for semantic modeling. (Representative models are shown on the right, w/ 5M to 15M parameters.)}
	\label{fig:paradigm}
\end{figure}

Despite significant progress in vision tasks, deep networks remain limited in modeling the workings of human visual system~{\cite{ref13,ref14,ref15}}. Unlike artificial neural network systems, the human visual system dynamically clusters scenes by grouping semantically consistent parts~{\cite{ref16,ref17,ref18}}. Conversely, convolutional and attention paradigms rely on fixed geometric windows (convolutional kernels or patches), constraining their ability to adaptively capture irregular objects in images~{\cite{ref19}}. Moreover, both paradigms prioritize performance through escalating architectural complexity, generating opaque abstractions that lack human interpretability ~{\cite{ref20}}.

To bridge this gap, the clustering paradigm~{\cite{ref16,ref19,ref21}} has emerged as a transparent alternative inspired by clustering algorithms.
This approach explicitly models images as sets of semantically coherent pixel groups, directly mirroring human perceptual organization.
Pioneering this direction, Ma \textit{et al.}~{\cite{ref21}} introduced Context-Cluster (CoC) for general vision tasks, leveraging unsupervised clustering at multiple feature levels to enable intrinsic interpretability via cluster visualization.
Building on this, Chen \textit{et al.}~{\cite{ref19}} proposed Feature Extraction with Clustering (FEC) which integrated clustering into pooling layers, enhancing model performance and enabling receptive field tracing for each feature point. ClusterFormer~{\cite{ref16}} incorporated cross-attention within iterative Expectation-Maximization (EM) steps to refine initial clusters solely from adjacency, enhancing cluster initialization and expanding spatial receptive fields.

Despite the enhanced model transparency offered by clustering paradigms, three key challenges persist: (1) Suboptimal performance: Current clustering-based models underperform relative to state-of-the-art vision architectures~{\cite{ref5,ref11}}, typically exceeding only pre-2022 baselines.
(2) Limited receptive field: Computational constraints force methods like CoC and ClusterFormer to perform local clustering within isolated windows. This design lacks inter-window communication, compromising semantic continuity and feature quality.
(3) Gradient vanishing in specialized components: FEC’s similarity projection layer in cluster pooling suffers from gradient vanishing on small datasets, freezing during training and causing redundancy and inefficiency.

In this paper, we propose a novel CLUster attEtion Network (CLUENet) to address the above discussed limitations.
The contributions of CLUENet are summarized as follows:
\begin{itemize}
	\item 
    Global Soft Aggregation and Hard Assignment allows CLUENet to compute global similarities between cluster centers and all pixels to form soft clusters via weighted fusion, while incorporating a gated residual module to supplement local context and a global query head to enable precise hard assignment for each pixel.
	
	\item 
    Efficient Aggregation with Shared Assignment employs cosine attention with learnable temperature in half-precision via FlashAttention for fast, memory-efficient aggregation, and shares hard assignment matrices across blocks within each stage to reduce redundancy and enhance stability.
	
	\item 
    Improved Cluster Pooling performs clustering and pooling in similarity space and projects results back to feature space via a perceptron, effectively alleviating gradient vanishing and enhancing performance.

    \item CLUENet achieves Top-1 accuracies of 76.55\% on CIFAR-100 and 82.44\% on Mini-ImageNet, outperforming existing clustering paradigm models and showing superior performance across paradigms.
\end{itemize}

\section{Related Work} \label{sec:related_work}

\textbf{Model interpretability.} Deep Neural Networks (DNNs) have achieved remarkable success in computer vision, yet their black-box nature raises concerns, especially in safety-critical domains such as medical diagnosis and autonomous driving~{\cite{ref27}}. Convolutional Neural Networks (CNNs) inherently lack intrinsic interpretability, making it difficult to intuitively reveal their decision rationale. Model-inspired deep architectures~\cite{huang2020learning,huang2022winnet,huang2025lightweight,pu2022mixed} improve model interpretability by embedding the formation model and priors into the model.
Existing visualization methods mainly focus on class activation maps~{\cite{ref28}}, feature visualization~{\cite{ref29}}, and input sensitivity analysis~{\cite{ref30}}. While these approaches improve interpretability to some extent, they are largely post-hoc and do not fully reveal the true internal decision mechanisms. With the rise of Vision Transformers (ViTs), attention-based interpretability has become a new research hotspot, including attention visualization~{\cite{ref8}} and emergent segmentation phenomena in self-supervised learning~{\cite{ref26}}. However, the interpretability of ViTs remains limited and uncertain. On the one hand, attention weights may be influenced by input noise or randomness during training, questioning their reliability as explanations~{\cite{ref31}}. On the other hand, most ViT interpretability still relies on post-hoc methods rather than uncovering internal decision processes. Recently, research focus has been shifting towards enhancing model transparency through intrinsically interpretable model architectures. Sparse Transformers~{\cite{ref32}} introduce sparse attention to highlight decision-critical features. Deep Nearest Centroids~{\cite{ref33}} offers analogy-based explanations by learning feature-to-centroid distances. And white-box Transformers like CRATE~{\cite{yu2023white}} are designed for observable and traceable attention paths and feature flows. 

\noindent\textbf{Clustering-based Model.}
The clustering paradigm proposed in 2023~{\cite{ref21}} innovatively applies the Simple Linear Iterative Clustering~{\cite{ref35}} to general visual representation tasks. This approach achieves strong performance in classification, detection, segmentation, and 3D point cloud processing while generating semantic grouping and visual explanations during inference, offering intuitive decision cues for users.
As the first systematic clustering paradigm model, CoC~{\cite{ref21}} represents images as spatially structured pixel groups and extracts features through a deep architecture inspired by a simplified clustering algorithm. 
Recently, ClusterFormer~{\cite{ref16}} integrates clustering with cross-attention in iterative EM steps. However, it relies on local windows, limiting its receptive field. 
FEC~{\cite{ref19}} supports global clustering but simplifies feature aggregation by averaging pixels per cluster without similarity weighting, reducing memory usage but risking gradient vanishing and limiting semantic complexity due to fewer clusters. Other recent works extend clustering and semantic aggregation to broader settings, highlighting the generality of cluster-based semantic modeling~\cite{ju2023glcc, liang2025from, guo2024depth}.
\begin{figure*}[!t]
	\centering
	\includegraphics[width=1.8\columnwidth]{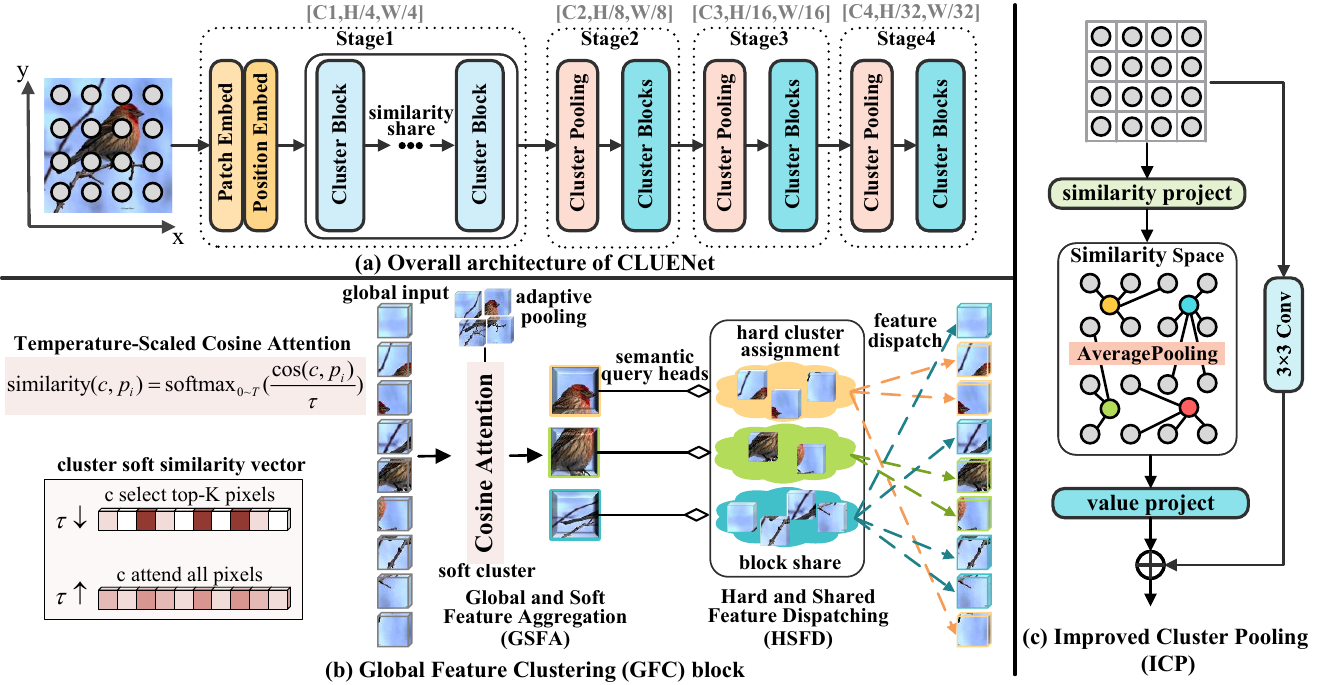}
	\caption{(a) Overall architecture of the CLUENet with a four-stage pyramid network; (b) The key components within the Global Feature Clustering (GFC) block, illustrating Global and Soft Feature Aggregation (GSFA) that updates cluster centers from all pixels, and Hard and Shared Feature Dispatching (HSFD) that updates pixel features according to their assigned cluster centers; (c) The Improved Cluster Pooling (ICP) block, depicting how pixel features are grouped into clusters in similarity space while preserving hierarchical structure.}
	\label{fig:methods}
\end{figure*}

\section{Proposed Method} \label{sec:proposed_method}

In this paper, we present CLUster attEntion Network (CLUENet), a transparent deep architecture for visual semantic understanding built on the clustering paradigm.
As shown in the overview of CLUENet in Fig.~\ref{fig:methods}(a), the proposed CLUENet uses a four-stage pyramid structure and consists of three novel components: (i) A Positional-aware Feature Embedding (PFE) Block encodes spatial relationships while preserving semantic locality.
(ii) Global Feature Clustering (GFC) Blocks (see Fig.~\ref{fig:methods}(b)) 
enable interpretable and flexible semantic modeling through global-soft aggregation paired with efficient hard dispatching, overcoming the fixed-window limitations.
(iii) Improved Cluster Pooling (ICP) Blocks (see Fig.~\ref{fig:methods}(c)) perform semantic information aggregation by grouping pixels into clusters in similarity space, maintaining hierarchical structure while mitigating gradient vanishing.
In the following, we introduce the key components of CLUENet in detail.

\subsection{Positional-aware Feature Embedding (PFE) Block}

For visual semantic clustering, positional information plays a crucial role in distinguishing structured visual regions. 
Conventional methods rely on fixed grid coordinates for positional embedding. However, this rigid approach lacks adaptability to scale variations. 
This limitation becomes particularly detrimental in scale-sensitive tasks such as instance and semantic segmentation~\cite{ref36}. 

To improve robustness to scale variations, we introduce a learnable convolutional positional embedding inspired by PosCNN~\cite{ref36} and PVTv2~\cite{ref11}. It employs lightweight depth-wise convolutions (DWConv) to encode spatial positional information. 
Formally, given an input image $\mathbf{I} \in \mathbb{R}^{H \times W  \times 3}$ and fixed grid coordinates $\mathbf{G} \in \mathbb{R}^{H \times W  \times 2}$ with $\mathbf{G}_{ij}=[\frac{i}{W}-0.5,\frac{j}{H}-0.5]$, the positional embedding is then performed as follows:
\begin{equation}
	\begin{aligned}
		\mathbf{X} &= \text{PatchEmbed}(\mathrm{Concat}[\mathbf{I}, \mathbf{G}]), \\
		\mathbf{X} &= \mathbf{X} +\text{DWConv}(\mathbf{X}),
	\end{aligned}
\end{equation}
where $\text{DWConv}(\cdot)$ denotes a depth-wise convolution operation with kernel size $k$, 
and $\text{PatchEmbed}$ refers to a non-overlapping convolutional layer with a patch size of $4 \times 4$.
The embedding is inserted after the initial patch embedding and within the feed-forward network of each attention block.

\begin{figure}[!t]
	\centering
	\includegraphics[width=\columnwidth]{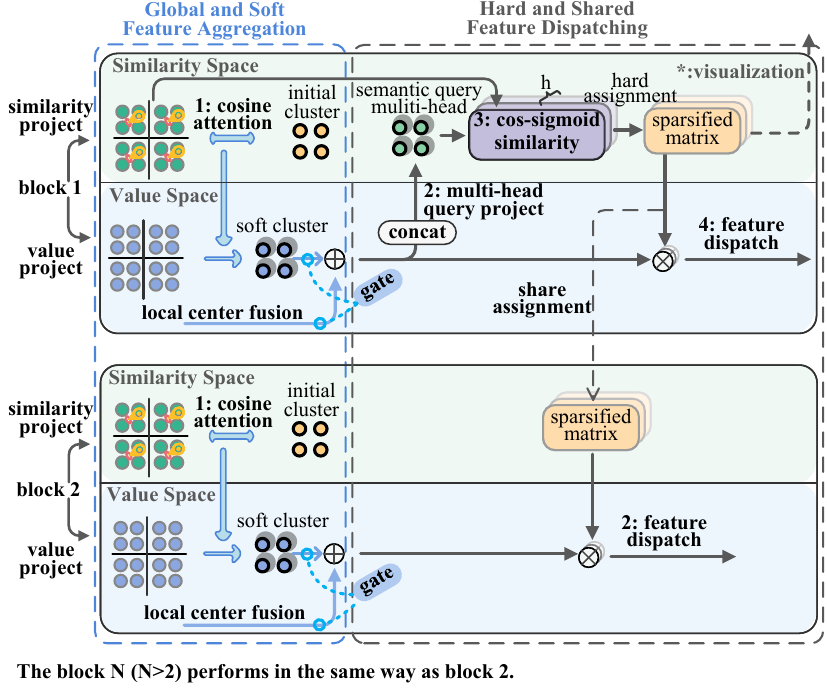}
	\caption{
    The details of the Global Feature Clustering (GFC) block. Global and Soft Feature Aggregation (GSFA) includes center initialization, Temperature-Scaled Cosine Attention, and Gated Fusion Mechanism. Hard and Shared Feature Dispatching (HSFD) includes multi-head query projection, hard clustering, and assignment shared across blocks.}
    
    
	\label{fig:details}
\end{figure}

\subsection{Global Feature Clustering (GFC) Block}
The GFC block serves as the core visual semantic clustering component in CLUENet.
As shown in Fig. \ref{fig:details}, the GFC block integrates a Global and Soft Feature Aggregation (GSFA) for cluster center updating and a Hard and Shared Feature Dispatching (HSFD) for pixel feature updating. They jointly enable the effective capture of semantic relationships while maintaining high computational efficiency. 

\subsubsection{Global and Soft Feature Aggregation (GSFA).}
Global aggregation enables each cluster center to dynamically attend to all pixels across the image, capturing long-range dependencies beyond local neighborhoods and promoting continuous clusters instead of disjoint window-based fragmented segments. 
And soft aggregation provides flexible, weighted attention over multiple pixels, enhancing feature aggregation and improving training stability.

As illustrated in Fig.~\ref{fig:details}, the input feature map $\mathbf{F} \in \mathbb{R}^{H \times W \times d}$ is first projected into similarity and value spaces via two linear transformations $\mathbf{W}_i \in \mathbb{R}^{d' \times1\times 1 \times d}$, producing feature maps $\mathbf{P}_i= \mathbf{W}_i \otimes \mathbf{F} \in \mathbb{R}^{H \times W \times d'}$ for $i \in \{s, v\}$.
The cluster centers $\mathbf{C}_i \in \mathbb{R}^{h \times w \times d'}$ are then obtained by applying 2D adaptive pooling on $\mathbf{P}_i$, partitioning the feature map into an $h \times w$ grid, yielding $m = h \times w$ cluster centers:
\begin{equation}
	\mathbf{C}_i = \operatorname{AdaptivePool}(\mathbf{P}_i, h, w).
\end{equation}

The cluster centers and pixel features are reshaped into 1D sequences $\mathbf{c}_i \in \mathbb{R}^{m \times d'}, \mathbf{p}_i \in \mathbb{R}^{n \times d'}, \text{where } n = H \times W.$

\textbf{Temperature-Scaled Cosine Attention: }
We propose a Temperature-Scaled Cosine Attention in which cosine similarity with a learnable temperature parameter and softmax operator is computed between cluster centers and all pixels, which is then used to perform weighted aggregation in the value space:
\begin{equation}
	\begin{aligned}
        &{{\mathbf{S}}_{C}} = {{\operatorname{softmax}}}\left( \frac{{\mathbf{c}}_{s} \cdot {{\mathbf{p}}}_{s}^{\top}}{ \tau \Vert \mathbf{c}_{s}\Vert \Vert \mathbf{p}_{s}\Vert} \right),   \\
		&\mathbf{c}{{'}_{v}} = {{\mathbf{S}}_{C}} \cdot {{\mathbf{p}_{v}}}, 
	\end{aligned}
\end{equation}
where ${{\mathbf{S}}}_{C}\in \mathbb{R}^{m\times n}$ is the similarity matrix between cluster centers and all pixels, 
$\mathbf{c}'_{v} \in \mathbb{R}^{m \times d'}$ represents the soft cluster centers obtained by weighted aggregation, and $\tau$ is a learnable temperature parameter.

This design allows each cluster center to retain similarity weights for all pixels across the global field while emphasizing differences in similarity, enabling selective and nuanced aggregation.

\textbf{Gated Fusion Mechanism: }
To flexibly incorporate local contextual information, we further introduce a gated fusion mechanism to replace the uniform local residual fusion used in prior work~{\cite{ref21}}. 
Specifically, grid-based local centers are first aggregated from the original features via grid partitioning, and then fed together with the soft cluster centers into the gating module:
\begin{equation}
	\begin{aligned}
		&\mathbf{g}=\sigma \left(f([\mathbf{c}_v,\mathbf{c}'_{v}])\right), \\ 
		&\widetilde{\mathbf{C}}_{v}=(1-\mathbf{g})\circ \mathbf{c}'_{v}+\mathbf{g}\circ \mathbf{c}_v,
	\end{aligned}
\end{equation}
where $\sigma(\cdot)$ is the sigmoid function, $f(\cdot)$ represents the gating network implemented as a two-layer perceptron, and $\mathbf{g}\in [0,1]^m$ is the gating weight vector which is broadcast along the feature dimension.


\subsubsection{Hard and Shared Feature Dispatching (HSFD).} We propose a hard and shared dispatching strategy to enable discrete and interpretable semantic modeling. Hard assignment enhances semantic distinctiveness by enforcing exclusive pixel-to-cluster associations, while shared dispatch rules across blocks within the same stage ensure consistency and reduce computational cost. 

The obtained cluster centers are projected into semantic query heads $\mathbf{q}\in {{\mathbb{R}}^{m\times d'}}$ (here we assume a single-head configuration for clarity) via an additional mapping layer.
The query head then computes the cosine similarities with pixel features ${{\mathbf{p}}_{s}}$ in the similarity space, followed by a sigmoid activation (referred to as cos-sigmoid in Fig.~\ref{fig:details}). 
\begin{equation}
	\mathbf{S}{_{P}}=\sigma \left( \alpha \left( \frac{{{\mathbf{p}}}_{s}\cdot {{\mathbf{q}}^{\top }}}{\Vert {\mathbf{p}}_{s} \Vert \Vert {\mathbf{q}} \Vert} \right)+\beta  \right),
\end{equation}
where $\alpha$ and $\beta$ are learnable scalars, initialized to 1 and 0, respectively.

A hard assignment strategy is then applied, where each pixel selects only its most similar cluster center. The resulting sparse similarity matrix $\widetilde{\mathbf{S}}_P$ is reused in the subsequent blocks:
\begin{equation}
	\begin{aligned}
		& {{\widetilde{\mathbf{S}}}_{P}}[i,j]=\mathbf{S}{_{P}}[i,j]\cdot \mathbf{1}\,(j=\underset{k}{\mathop{\arg \max }}\,\mathbf{S}{_{P}}[i,k]), \\ 
		&\mathbf{{P}'}=\mathbf{P}+\text{FC}\left( {{\widetilde{\mathbf{S}}}_{P}}\cdot {{\widetilde{\mathbf{C}}}_{v}} \right),
	\end{aligned}
\end{equation}
where $\mathbf{S}_P \in \mathbb{R}^{n \times m}$ denotes the similarity between each pixel and the query heads, $\widetilde{\mathbf{S}}_P$ is the sparsified matrix after hard assignment, and FC($\cdot$) denotes a fully connected layer used to project features from hidden dimension $d'$ back to the original dimension $d$. 

\textbf{Multi-Head Clustering: }
Inspired by multi-head attention~{\cite{ref23}}, we split semantic queries and pixel features along the feature dimension into $M$ heads, enabling parallel cluster assignment and improving semantic modeling accuracy. Specifically, $\mathbf{p}_i$ is first partitioned into $M$ heads $\mathbf{p}_i = \{\mathbf{p}_i^{(k)} \in \mathbb{R}^{n \times d'/M} \}_{k=1}^M, i\in \{s,v\}$.
These multi-head pixel features induce multi-head cluster centers in the value space $\widetilde{\mathbf{C}}_v = \{\widetilde{\mathbf{C}}_v^{(k)} \in \mathbb{R}^{m \times d'/M}\}_{k=1}^M$.
Then, the multi-head cluster centers $\widetilde{\mathbf{C}}_v$ are concatenated along the channel dimension and linearly projected via a learnable matrix $\mathbf{W}_q \in \mathbb{R}^{d' \times d'}$ to form query features $\mathbf{q}' = \mathrm{concat}\left(\widetilde{\mathbf{C}}_v^{(1)},  \ldots, \widetilde{\mathbf{C}}_v^{(M)}\right) \cdot \mathbf{W}_q \in \mathbb{R}^{m \times d'}$.
The projected queries $\mathbf{q}'$ are also split back into $M$ heads $\mathbf{q} = \{\mathbf{q}^{(k)} \in \mathbb{R}^{m \times d'/M}\}_{k=1}^M$ which serve as semantic query heads for the hard assignment in each subspace.

\subsection{Improved Cluster Pooling (ICP) Block}
\begin{figure}[!t]
\centering
\includegraphics[width=\columnwidth]{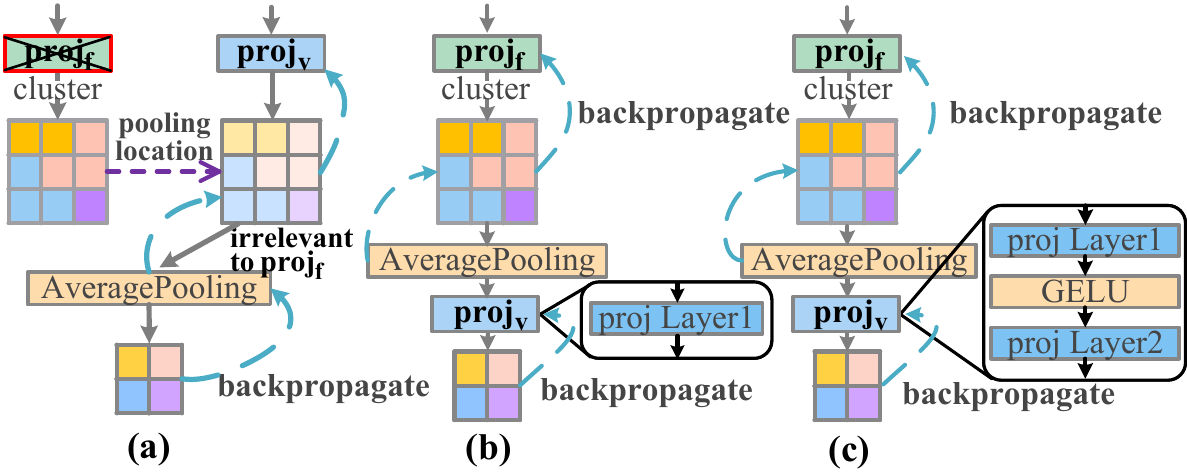}
\caption{Cluster pooling configurations. 
(a) In FEC, $\operatorname{proj_f}$ only guides pixel selection and does not update, (b) connecting $\operatorname{proj_f}$ and $\operatorname{proj_v}$ avoids gradient issues but offers limited performance gains, (c) The proposed cluster pooling adopts a two-layer perceptron for $\operatorname{proj_v}$, enabling effective training and improved performance.}
\label{fig:ClusterPool}
\end{figure}

We identify a key limitation in the cluster pooling design of FEC~\cite{ref19}. As shown in Fig.~\ref{fig:ClusterPool}, the projection layer $\operatorname{proj_f}$ which maps features into the similarity space, merely marks pixel-cluster associations during the forward pass but does not participate in the actual aggregation. This leads to a disconnection between the forward data flow and the learnable parameters of $\operatorname{proj_f}$, resulting in ineffective gradient backpropagation and a near-identity behavior during training. 

To address this, we propose to perform average pooling directly in the similarity space, making the aggregation explicitly dependent on $\operatorname{proj_f}$’s output and ensuring effective gradient updates. Meanwhile, the $\operatorname{proj_v}$ mapping layer is redefined to map the aggregated similarity information back to the feature space. Notably, we adopt a two-layer perceptron for $\operatorname{proj_v}$ instead of a linear layer, as experiments show that single-layer designs offer limited performance gains. With these modifications, both $\operatorname{proj_f}$ and $\operatorname{proj_v}$ are effectively optimized during training, leading to significant performance improvements.

\section{Experiment} \label{sec:experiment}
This section presents a comprehensive evaluation of the proposed CLUENet, conducts visualization analysis and ablation studies\footnote{Further experimental details, visualizations, and failure case analyses are available in the supplementary material.}.

\subsection{Experiment Settings}

\setlength{\tabcolsep}{2pt} 
\renewcommand{\arraystretch}{1.1} 
\begin{table}[!t]
\small
\centering
\begin{tabular}{
>{\centering\arraybackslash}m{0.1cm} 
>{\centering\arraybackslash}p{2.2cm}
>{\centering\arraybackslash}p{1.0cm}
>{\centering\arraybackslash}p{1.0cm}
>{\centering\arraybackslash}p{0.9cm}
>{\centering\arraybackslash}p{1.0cm}
>{\centering\arraybackslash}p{1.1cm}}
\toprule
\multicolumn{2}{c}{\textbf{Method}} & \textbf{\#Param} & \textbf{FLOPs} & \textbf{Top-1} & \textbf{FPS} & \textbf{Memory} \\
\midrule
\multirow{4}{*}{\rotatebox[origin=c]{90}{\textbf{Convolution}}} 
& ResNet & 14.16 & 186.21 & 69.84 & 16858\raisebox{-0.20ex}{\scriptsize ±431} & 1.1 \\
& ConvMixer & 2.78 & 175.88 & 67.21 & \textbf{20433}\raisebox{-0.20ex}{\scriptsize ±908} & \textbf{0.6} \\
& ShuffleNet & 5.55 & 186.83 & \textbf{71.09} & 14660\raisebox{-0.20ex}{\scriptsize ±290} & 1.0 \\
& MobileNet & 2.37 & 68.43 & 66.62 & 17241\raisebox{-0.20ex}{\scriptsize ±862} & 0.7 \\
\midrule
\multirow{4}{*}{\rotatebox[origin=c]{90}{\textbf{Attention}}} 
& ViT & 3.22 & 224.12 & 56.93 & \textbf{16809}\raisebox{-0.20ex}{\scriptsize ±570} & \textbf{0.8} \\
& PVTv2 & 3.43 & 172.88 & \textbf{70.77} & 12711\raisebox{-0.20ex}{\scriptsize ±158}  & 1.2 \\
& CPVT & 3.12 & 155.27 & 66.09 & 14885\raisebox{-0.20ex}{\scriptsize ±215} & 1.2 \\
& Swin & 5.15 & 245.84 & 65.33 & 11496\raisebox{-0.20ex}{\scriptsize ±181} & 1.6 \\
\midrule
\multirow{4}{*}{\rotatebox[origin=c]{90}{\textbf{Cluster}}} 
& CoC & 2.72 & 161.11 & 71.92 & \textbf{10712}\raisebox{-0.20ex}{\scriptsize ±98} & \textbf{1.4} \\
& FEC & 2.83 & 197.12 & 69.73 & 9663\raisebox{-0.20ex}{\scriptsize ±362} & 1.4 \\
& ClusterFormer & 2.92 & 173.47 & 66.05 & 8041\raisebox{-0.20ex}{\scriptsize ±46} & 1.6 \\
& \textbf{CLUENet} (ours) & 3.02 & 188.88 & \textbf{76.55} & 7807\raisebox{-0.20ex}{\scriptsize ±280} & 1.4 \\
\bottomrule
\end{tabular}
\caption{Comparison with representative backbones on CIFAR-100 benchmark.}
\label{tab:cifar_results}
\end{table}

\setlength{\tabcolsep}{2pt} 
\renewcommand{\arraystretch}{1.1} 
\begin{table*}[!t]
	\small
	\centering
	\begin{tabular}{
			>{\centering\arraybackslash}p{0.6cm} 
			>{\centering\arraybackslash}p{5.2cm} 
			>{\centering\arraybackslash}p{1.2cm}
			>{\centering\arraybackslash}p{1.2cm}
			>{\centering\arraybackslash}p{1.2cm}
			>{\centering\arraybackslash}p{1.2cm}
			>{\centering\arraybackslash}p{1.5cm}
			>{\centering\arraybackslash}p{1.2cm}
		}
		\toprule
		\multicolumn{2}{c}{\textbf{Method}} & \textbf{\#Param} & \textbf{FLOPs} & \textbf{Top-1} & \textbf{Top-3} & \textbf{FPS} & \textbf{Memory} \\
		\midrule
		\multirow{6}{*}{\rotatebox[origin=c]{90}{\makecell{\textbf{Convolution-based} \\ \textbf{method}}}} 
		& ResNet18~{\cite{ref3}} & 14.17 & 2.38 & 76.95 & 89.88 & 1150.75\raisebox{-0.20ex}{\scriptsize ±27.15} & 2.7 \\[1pt]
		& ShuffleNetv2 (x1.5)~{\cite{ref6}} & 2.58 & 0.31 & \textbf{78.39} & \textbf{90.40} & \textbf{1194.72}\raisebox{-0.20ex}{\scriptsize ±23.26} & \textbf{1.1} \\[1pt]
		& ShuffleNetv2 (x2.0)~{\cite{ref6}} & 6.52 & 0.64 & \textbf{79.63} & \textbf{90.93} & \textbf{1205.17}\raisebox{-0.20ex}{\scriptsize ±7.59} & \textbf{1.3} \\[1pt]
		& ConvNeXtv2 (A)~{\cite{ref43}} & 3.42 & 0.55 & 71.15 & 84.97 & 1172.04\raisebox{-0.20ex}{\scriptsize ±34.80} & 1.6 \\[1pt]
		& ConvNeXtv2 (F)~{\cite{ref43}} & 4.89 & 0.79 & 73.14 & 86.33 & 1177.66\raisebox{-0.20ex}{\scriptsize ±21.15} & 1.8 \\[1pt]
		& ConvNeXtv2 (N)~{\cite{ref43}} & 15.05 & 2.46 & 75.04 & 87.48 & 1148.19\raisebox{-0.20ex}{\scriptsize ±28.45} & 2.8 \\
        \midrule
		\multirow{8}{*}{\rotatebox[origin=c]{90}{\makecell{\textbf{Attention-based} \\ \textbf{method}}}} 
		& PVTv2 (b0)~{\cite{ref11}} & 3.44 & 0.54 & 75.34 & 88.52 & 1195.32\raisebox{-0.20ex}{\scriptsize ±31.63} & 1.7 \\
		& PVTv2 (b1)~{\cite{ref11}} & 13.55 & 2.06 & 77.57 & 89.86 & 1154.58\raisebox{-0.20ex}{\scriptsize ±39.86} & 2.9 \\
		& EfficientFormer (s0)~{\cite{ref44}} & 3.26 & 0.40 & \textbf{77.92} & \textbf{89.71} & 1166.09\raisebox{-0.20ex}{\scriptsize ±21.46} & 1.5 \\
		& EfficientFormer (s1)~{\cite{ref44}} & 5.76 & 0.66 & \textbf{78.97} & \textbf{90.32} & 1128.59\raisebox{-0.20ex}{\scriptsize ±0.75} & 1.5 \\
		& EfficientFormer (s2)~{\cite{ref44}} & 12.16 & 1.27 & \textbf{79.75} & \textbf{90.42} & 695.41\raisebox{-0.20ex}{\scriptsize ±1.71} & 1.5 \\
		& EfficientViT (m2)~{\cite{ref45}} & 3.99 & 0.20 & 73.59 & 86.78 & \textbf{1173.13}\raisebox{-0.20ex}{\scriptsize ±22.48} & \textbf{0.9} \\
		& EfficientViT (m3)~{\cite{ref45}} & 6.61 & 0.26 & 73.88 & 87.16 & \textbf{1132.69}\raisebox{-0.20ex}{\scriptsize ±24.25} & \textbf{1.0} \\
		& EfficientViT (m5)~{\cite{ref45}} & 12.13 & 0.52 & 75.32 & 88.03 & \textbf{1160.99}\raisebox{-0.20ex}{\scriptsize ±44.32} & \textbf{1.0} \\
		\midrule
		\multirow{13}{*}{\rotatebox[origin=c]{90}{\textbf{Clustering-based method}}} 
		& CoC (tiny)~{\cite{ref21}} & 5.28 & 1.12 & 74.90 & 88.29 & \textbf{887.11}\raisebox{-0.20ex}{\scriptsize ±18.92} & \textbf{2.2} \\
		& CoC (small)~{\cite{ref21}} & 14.20 & 2.80 & 76.56 & 89.13 & \textbf{883.46}\raisebox{-0.20ex}{\scriptsize ±29.31} & \textbf{3.2} \\
            & CoC (medium)~{\cite{ref21}} & 28.83 & 5.96 & 78.39 & 90.29 & 605.67\raisebox{-0.20ex}{\scriptsize ±3.18} & \textbf{4.2} \\
		& ClusterFormer~{\cite{ref16}}$^{\dagger}$ & 5.63 & 1.24 & 70.73 & 85.73 & 669.00\raisebox{-0.20ex}{\scriptsize ±4.52} & 9.3 \\
		& ClusterFormer~{\cite{ref16}}$^{\dagger}$ & 15.01 & 3.02 & 73.93 & 87.29 & 694.91\raisebox{-0.20ex}{\scriptsize ±5.99} & 12.0 \\
        & ClusterFormer (tiny)~{\cite{ref16}} & 30.27 & 5.58 & 73.99 & 88.26 & 588.56\raisebox{-0.20ex}{\scriptsize ±5.11} & 9.6 \\
		& FEC (small)~{\cite{ref19}} & 5.44 & 1.38 & 76.74 & 89.34 & 854.25\raisebox{-0.20ex}{\scriptsize ±37.49} & 5.2 \\
		& FEC (base)~{\cite{ref19}} & 14.63 & 3.37 & 77.81 & 90.26 & 859.63\raisebox{-0.20ex}{\scriptsize ±16.23} & 5.3 \\
            & FEC (large)~{\cite{ref19}} & 29.26 & 6.55 & 79.33 & 90.55 & 550.69\raisebox{-0.20ex}{\scriptsize ±2.58} & 7.6 \\
		& \textbf{CLUENet (micro)} & 3.02 & 0.65 & \textbf{78.75} & \textbf{90.83} & 871.89\raisebox{-0.20ex}{\scriptsize ±13.84} & 4.2 \\
		& \textbf{CLUENet (tiny)} & 5.68 & 1.30 & \textbf{80.51} & \textbf{91.29} &  867.76\raisebox{-0.20ex}{\scriptsize ±17.89} & 5.5 \\
		& \textbf{CLUENet (small)} & 15.05 & 3.16 & \textbf{81.49} & \textbf{92.06} & 878.60\raisebox{-0.20ex}{\scriptsize ±21.68} & 6.5 \\
        & \textbf{CLUENet (base)} & 30.20 & 6.40 & \textbf{82.44} & \textbf{92.47} & \textbf{679.05}\raisebox{-0.20ex}{\scriptsize ±7.12} & 7.7 \\
		\bottomrule
	\end{tabular}
	\caption{Comparison with representative backbones on Mini-ImageNet benchmark. ($^{\dagger}$ indicates architectures adjusted to reach the specified parameter scale.)}
	\label{tab:miniimagenet_results}
	\vspace{1mm}
	\noindent
	\begin{minipage}[t]{\dimexpr\linewidth-5.2cm\relax} 
		\footnotesize
		\raggedright
		\setlength{\baselineskip}{10pt}
	\end{minipage}
\end{table*}

\textbf{Datasets.} 
CIFAR-100 and Mini-ImageNet are used for the experiment.
CIFAR-100 contains 50K training and 10K validation images of size $32 \times 32$. Mini-ImageNet is a subset of ImageNet-1K with 100 classes and 600 images per class. We split Mini-ImageNet 4:1 into training and validation sets, ensuring balanced class distribution. 

We follow the standard data augmentation procedure during training. For CIFAR-100, random horizontal flip and normalization are applied. For Mini-ImageNet, random crop, horizontal flip, color jitter, random erasing, and normalization are applied for training.

\noindent\textbf{Training Details.}
All models are trained using the AdamW optimizer with momentum 0.9 and cosine decay. Weight decay is set to 0.05. The learning rate is auto-tuned using \texttt{findLR}~\cite{ref42}, with a warmup of 5 epochs. The batch size for CIFAR-100 is 256, and the batch size for Mini-ImageNet is 128. All experiments are conducted on a computer with an NVIDIA RTX 4090 (24GB) GPU. 


\noindent\textbf{Evaluation Protocol.}
Following common practice~\cite{ref16,ref19,ref21}, all models are evaluated on the validation set after completing 100 training epochs, rather than using the model with the best validation performance, to ensure a fair comparison. Each model is evaluated 7 times; the first 2 runs are discarded to avoid cold-start effects. We report the average of the last 5 runs for classification accuracy (Top-1 and Top-3), throughput (Frames Per Second, FPS), number of parameters (\#Param), FLOPs, and inference memory usage, while all metrics except FPS show no significant change.

\subsection{Image Classification Results}

\textbf{Results on CIFAR-100.} 
As shown in Table~\ref{tab:cifar_results}, in the convolution paradigm, ShuffleNetv2 achieves the best Top-1 accuracy of 71.09\%. For the attention paradigm, PVTv2 leads with 70.77\%. Among clustering paradigm, CLUENet stands out with a Top-1 accuracy of 76.55\%, surpassing existing clustering model CoC by 4.63\% and outperforming PVTv2 and ShuffleNetv2 by 5.78\% and 5.46\%, respectively. 

\noindent\textbf{Results on Mini-ImageNet.}
Table~\ref{tab:miniimagenet_results} summarizes the classification results on the Mini-ImageNet validation set. Under similar parameter scales, CLUENet (base) achieves the best performance among clustering-based models, surpassing CoC (medium), FEC (large), and ClusterFormer (tiny) by 4.05\%, 3.11\%, and 8.45\% in Top-1 accuracy, and by 2.18\%, 1.92\%, and 4.21\% in Top-3 accuracy, respectively. 
Without using windowing mechanisms while adopting a relatively dense cluster count, CLUENet (base) achieves an inference speed of 679.05 img/s, which is comparable to that of FEC (550.69 img/s) and CoC (605.67 img/s), demonstrating its efficient architecture and strong feature representation.

Compared with the Attention-based models, CLUENet (micro) outperforms the attention-based PVTv2 (b0) by 3.41\% in Top-1 accuracy and 2.31\% in Top-3 accuracy while using 0.42M fewer parameters. It also surpasses EfficientFormer (s0) by 0.83\% in Top-1 and 1.12\% in Top-3 accuracy with a similar parameter budget. CLUENet (tiny) outperforms EfficientFormer (s1) by 1.54\% in Top-1 and 0.97\% in Top-3 accuracy. Furthermore, CLUENet (small) surpasses PVTv2 (b1) and EfficientFormer (s2) by 3.92\% and 1.74\% in Top-1, respectively.

Compared with the convolution-based models, CLUENet (micro) achieves slightly higher Top-1 and Top-3 accuracy than ShuffleNetv2 (x1.5) (+0.36\%, +0.43\%), and CLUENet (tiny) outperforms ShuffleNetv2 (x2.0) and ResNet18 by 0.88\% and 3.56\% in Top-1 accuracy, and by 0.36\% and 1.41\% in Top-3 accuracy, respectively. Moreover, CLUENet (small) surpasses both ResNet18 and ConvNeXtv2 (N) by 4.54\% and 6.45\% in Top-1 accuracy, respectively.

\subsection{Visualization Analysis}

\begin{figure*}[t]
    \centering
    
    \hfill
    \begin{subfigure}[t]{0.45\linewidth} 
        \centering
        \includegraphics[width=\linewidth]{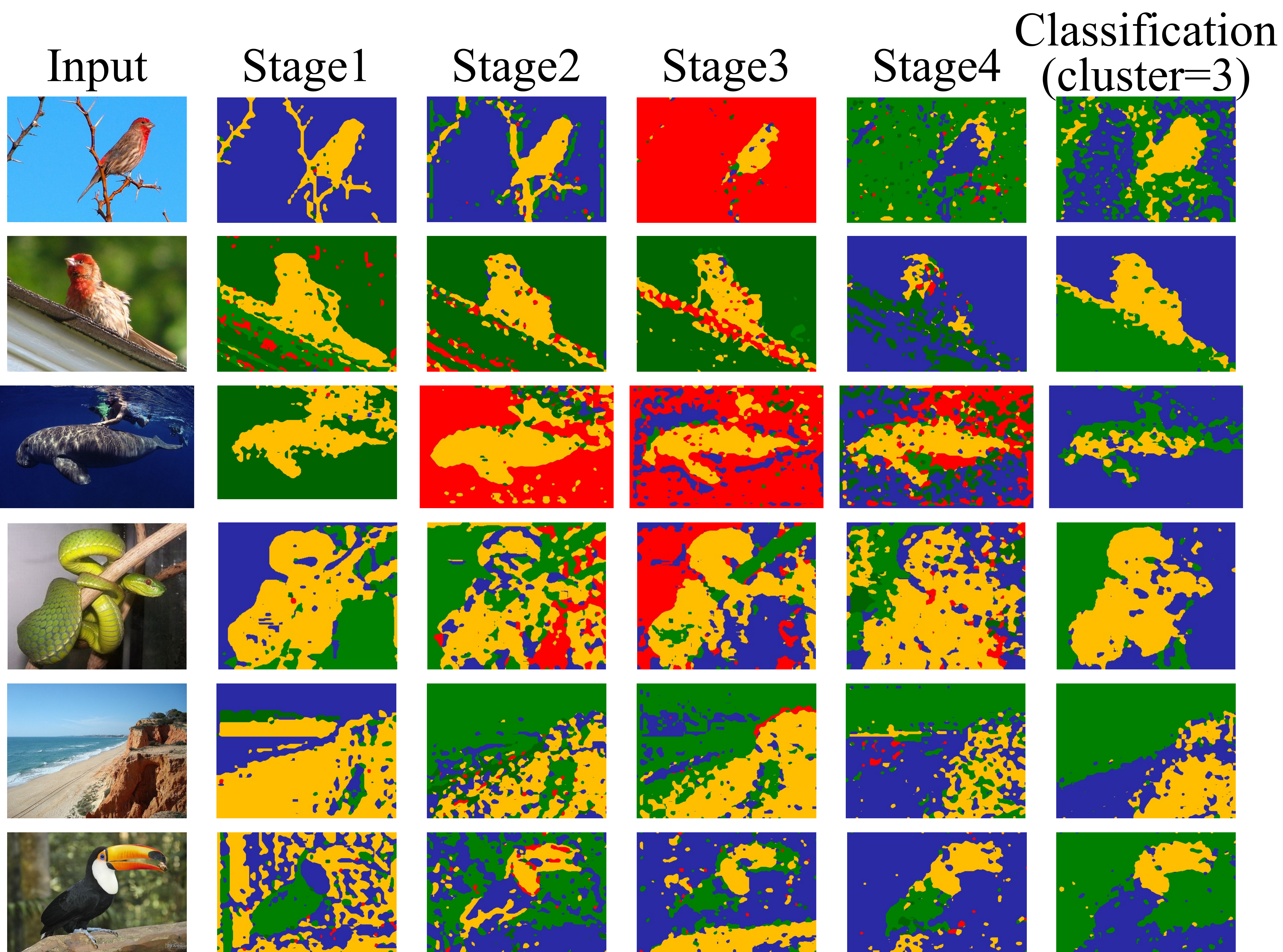} 
        \caption{The semantic clustering.} 
        \label{fig:vis_miniimagenet_a}
    \end{subfigure}
    \hfill 
    \begin{subfigure}[t]{0.45\linewidth}
        \centering
        \includegraphics[width=\linewidth]{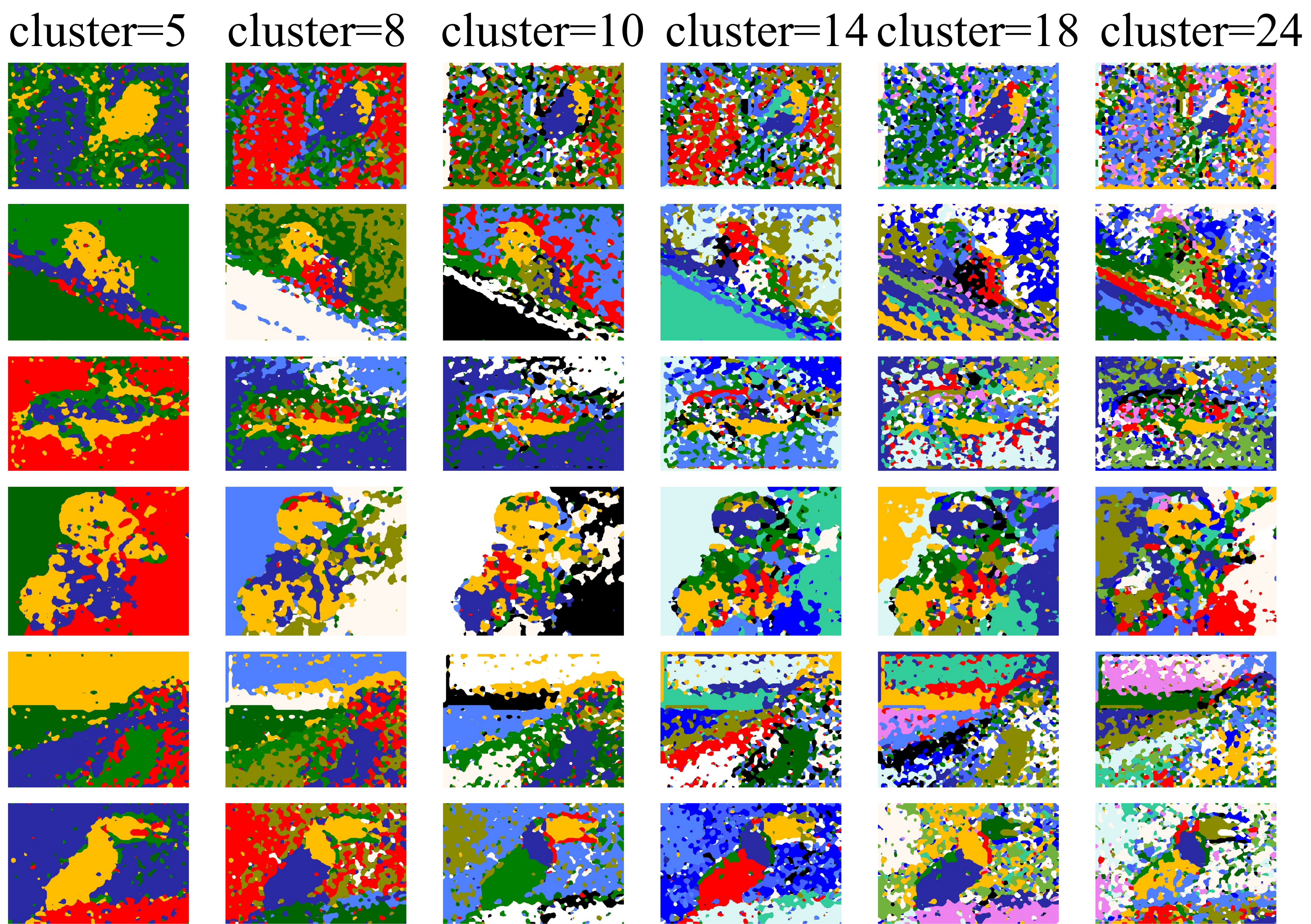}
        \caption{The impact of cluster numbers.}
        \label{fig:vis_miniimagenet_b}
    \end{subfigure}
    \hfill
    
    \caption{ Visualization of (a) the clustering results of semantic heads at each of the four stages, along with the global receptive field map \textit{w.r.t.} the final classification decision, and (b) the global receptive field map \textit{w.r.t.} different cluster numbers.}
    \label{fig:vis_miniimagenet}
\end{figure*}

\textbf{Model Visualization. }The proposed CLUENet offers intrinsic interpretability through its architecture. 
We visualize the cluster receptive fields of CLUENet (tiny) across all four stages on the Mini-ImageNet classification task. As shown in Fig.~\ref{fig:vis_miniimagenet}, the model captures multi-level semantic structures at different stages, reflecting a hierarchical feature extraction process similar to human visual perception. For clarity, only one semantic head is visualized per stage. Additionally, we visualize the receptive field of each position on the $7\times7$ feature map from the fourth stage to highlight the semantic areas most influential for classification. Since directly displaying all 49 clusters can be overwhelming, we adopt the $K$-Means-based cluster merging method proposed by Chen \textit{et al.}~{\cite{ref19}} to reduce the number of visualized clusters. To preserve the interpretability purity of the network, we visualize the fourth-stage $7\times7$ feature map with increasing cluster numbers by adjusting the K-Means merge parameter. 

\begin{figure}[!t]
	\centering
	\includegraphics[width=\columnwidth]{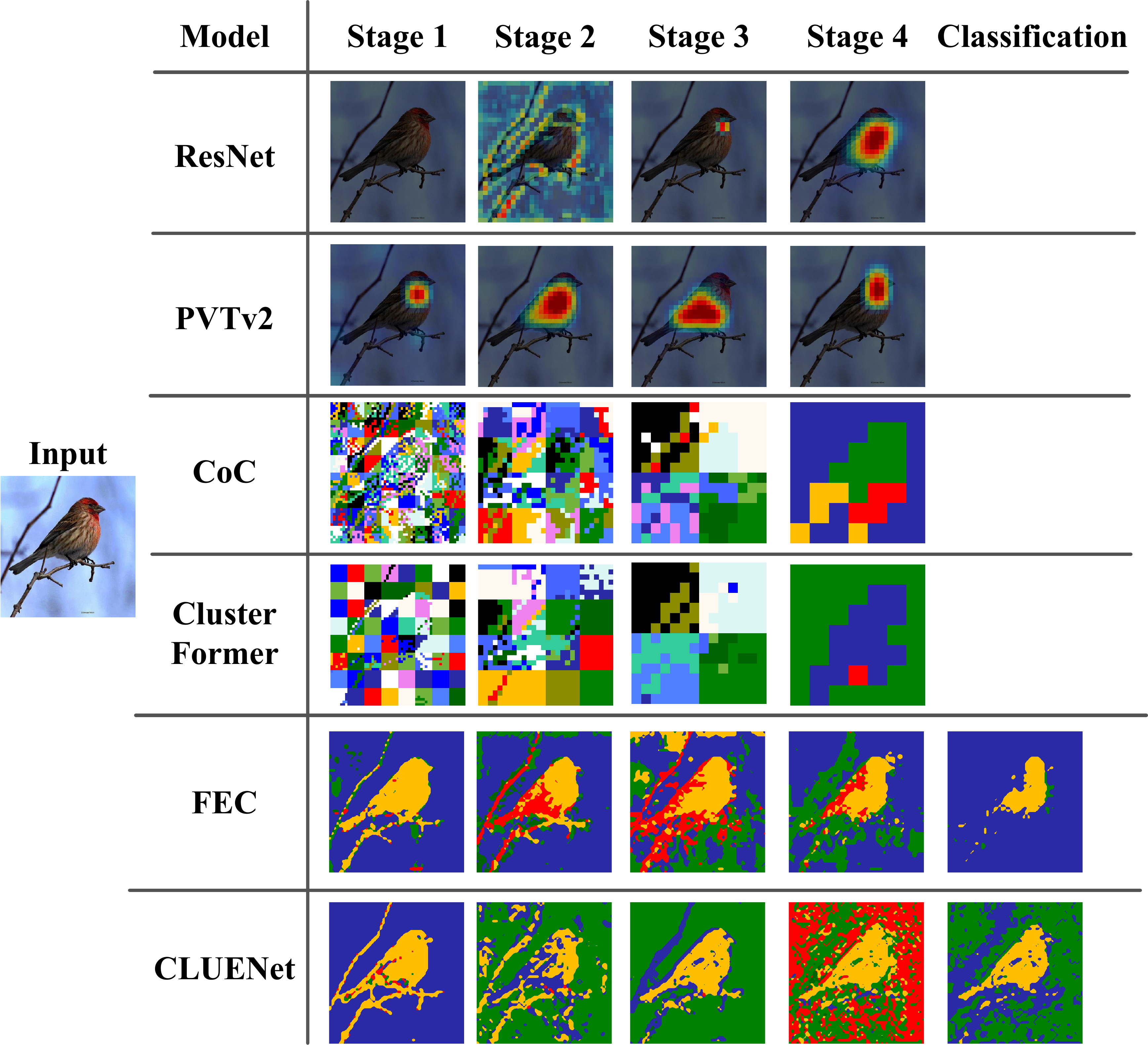}
	\caption{Visualization results of different paradigm models. ResNet shows Grad-CAM activation maps at four stages. PVTv2 presents attention maps based on the center query token in the first attention head of the last block at each stage. CoC, ClusterFormer and FEC visualize one attention head from one block per stage. FEC and CLUENet additionally provide the final-stage semantic clustering maps used for classification.}
	\label{fig:visual_compare}
\end{figure}

As illustrated in Fig.~\ref{fig:vis_miniimagenet_a}, CLUENet exhibits a progressive semantic understanding. Early stages (1–3) focus on the object’s overall contour and location, separating it from the background to form clear object boundaries. The deeper stages (3–4) progressively shift attention to finer and more discriminative regions critical for classification. For example, in the first row of Fig.~\ref{fig:vis_miniimagenet_a}, the model initially treats the bird and branch as a whole, but by Stage 3, it distinguishes the branch as background and focuses on the bird’s head and torso in Stage 4. 
As shown in Fig.~\ref{fig:vis_miniimagenet_b}, when the number of clusters is further increased, the same semantic object is split into multiple semantic parts, capturing finer details of the object.
Importantly, all visualized samples are correctly classified, and the final-stage feature maps (the last column of Fig.~\ref{fig:vis_miniimagenet_a}) consistently exhibit clear cluster structures that concentrate on the target object and effectively distinguish it from the background.

\noindent\textbf{Cross-Paradigm Visualization Comparison. } Fig.~\ref{fig:visual_compare} presents a comparative visualization analysis of different paradigms on the Mini-ImageNet classification task. For clustering-based models (CoC, ClusterFormer, FEC and CLUENet), we directly visualize the semantic clusters at each stage. For FEC and our proposed CLUENet, we additionally show the final-stage clustering maps to illustrate their global semantic representations. For convolution-based models (\textit{e.g.}, ResNet), we use Grad-CAM~{\cite{ref29}} to highlight class-discriminative regions. For attention-based models (\textit{e.g.}, PVTv2), we visualize the attention response to the center query in the first attention head of the last block at each stage. It is observed that ResNet converges on the head, while PVTv2 shifts focus from the torso to the head. Both CoC and ClusterFormer produce fragmented clusters due to local window constraints, thereby lacking coherent global semantics. In contrast, FEC and CLUENet consistently exhibit clear foreground-background separation across stages, indicating stronger interpretability. Moreover, CLUENet combines high accuracy, efficiency, and transparency, surpassing FEC in overall performance.

\begin{table*}[t]
\footnotesize
\centering
\begin{minipage}[b]{0.48\linewidth}
\centering
\begin{tabular}{ccccc|cc|cc}
\toprule
FA & TCosAttn & Gate & Shared & PosEmb & Param & FLOPs  & Top-1 & Top-3 \\
\midrule
\xmark & \xmark & \xmark & \cmark & \cmark & 2.87 & 0.518  & 76.97 & 89.62 \\
\cmark & \xmark & \xmark & \cmark & \cmark & 2.98 & 0.649  & 77.87 & 90.27 \\
\cmark & \cmark & \xmark & \cmark & \cmark & 2.98 & 0.649  & 78.46 & 90.38 \\
\cmark & \cmark & \cmark & \xmark & \cmark & 3.12 & 0.663  & 78.32 & 90.05 \\
\cmark & \cmark & \cmark & \cmark & \xmark & 2.97 & 0.621  & 78.21 & 90.17 \\
\cmark & \cmark & \cmark & \cmark & \cmark & 3.02 & 0.649  & \textbf{78.75} & \textbf{90.83} \\
\bottomrule
\end{tabular}
\\[2pt]
(a) Key Component  
\end{minipage}
\hfill
\begin{minipage}[b]{0.48\linewidth}
\centering
\begin{tabular}{ll|cccc}
\toprule
\multicolumn{2}{c|}{Design} & Top-1 & Top-3 & FPS & FLOPs \\
\midrule
\multicolumn{2}{c|}{w/o Cluster Pooling} & 77.38 & 89.78 & 899.03 & 0.539 \\
\midrule
\multirow[c]{2}{*}{Parallel} & FEC  & 78.07 & 89.96 & 846.61 & 0.769 \\
& FEC ($\operatorname{proj_f}$ = identity) & 78.09 & 89.96 & 893.92 & 0.635 \\
\midrule
\multirow[c]{3}{*}{Sequential} & Ours (Single Layer) & 78.10 & 90.07 & 879.83 & 0.641 \\
& {\textbf{Ours (2-layer MLP)}} & {\textbf{78.75}} & {\textbf{90.83}} & 871.89 & 0.649 \\
& Ours (3-layer MLP) & 78.71 & 90.41 & 856.19 & 0.667 \\
\bottomrule
\end{tabular}
\vspace{2pt}  
\par(b) Cluster Pooling Configurations 
\end{minipage}

\vspace{1em}

\begin{minipage}[b]{0.48\linewidth}
\centering
\begin{tabular}{l|cc}
\toprule
Design & Top-1 & Top-3 \\
\midrule
w/o Query & 77.72 & 90.06  \\
AvgPool Query & 78.02 & 90.11 \\
Single-head center Query & 78.41 & 90.27  \\
{\textbf{Multi-head center Query}} & {\textbf{78.75}} & {\textbf{90.83}}  \\
\bottomrule
\end{tabular}
\vspace{2pt}  
\par(c) Multi-Head Semantic Query 
\end{minipage}
\hfill
\begin{minipage}[b]{0.48\linewidth}
\centering
\begin{tabular}{l|ccccc}
\toprule
Design & FLOPs & Mem & FPS & Top-1 & Top-3 \\
\midrule
+Fewer Clusters ($\le$16) & 0.637 & 4.1 & 882.63 & 78.06 & 90.18 \\
+Window Partition  & 0.636 & 4.1 & 881.40 & 78.30 & 90.37 \\
{Ours ($64$)}  & 0.653 & 4.2 & 870.36 & 78.70 & 90.62 \\
{\textbf{Ours} ($49$)}  & 0.649 & 4.2 & 871.89 & {\textbf{78.75}} & {\textbf{90.83}} \\
\bottomrule
\end{tabular}
\vspace{2pt}  
\par(d) Model Design Comparison 
\end{minipage}
\caption{Ablation Studies of Core Components, Cluster Pooling, Query Design, and Model Variants}
\label{tab:ablation_combined}
\end{table*}

\subsection{Ablation Study}

We validate the effectiveness of model components on the Mini-ImageNet image classification task. 

\noindent\textbf{Key Components:} As shown in Table~\ref{tab:ablation_combined}(a), we conducted ablation studies on five core components of the model to evaluate their individual contributions. Specifically, these components include: Feature Aggregation (FA), Temperature-Scaled Cosine Attention (TCosAttn), Gated Residual (Gate), Shared Dispatching (Shared), and Learnable Positional Embedding (PosEmb). 
Since both TCosAttn and Gated Residual are embedded within the global and soft feature aggregation module, removing FA also disables the other two. Disabling FA leads to a significant drop in accuracy (Top-1: –1.78\%, Top-3: –1.21\%), highlighting the importance of global and soft feature aggregation in semantic integration. 
Replacing TCosAttn with standard attention reduces Top-1 and Top-3 accuracy by 0.59\% and 0.11\%, respectively, indicating its advantage in modeling semantic hierarchy and improving cluster-to-pixel alignment. Removing the Gated Residual and using fixed weights results in a smaller drop (–0.29\%, –0.45\%), showing that dynamic gating better adapts to local feature injection. Disabling shared dispatching causes accuracy to decrease by 0.43\% (Top-1) and 0.78\% (Top-3), along with increased parameters and computation. Finally, removing the learnable positional embedding leads to a 0.54\% drop in Top-1 and 0.66\% in Top-3 accuracy, suggesting its role in enhancing spatial awareness. 
Overall, all five components contribute meaningfully to performance and work together to improve the model’s representation capacity.

\noindent\textbf{Cluster Pooling:} Table~\ref{tab:ablation_combined}(b) compares various cluster pooling designs. The similarity mapping layer $\operatorname{proj_f}$ in FEC shows performance nearly identical to an identity mapping, indicating model redundancy. Even when both $\operatorname{proj_f}$ and $\operatorname{proj_v}$ layers learn jointly (Fig.~\ref{fig:ClusterPool}b), accuracy gains are negligible. Our improved cluster pooling module achieves a notable Top-1 accuracy gain (78.75\% vs. 78.07\%) while reducing computational cost, validating the design’s efficiency. Increasing $\operatorname{proj_v}$ depth from two to three layers yields no improvement, thus, the two-layer MLP is selected for balance.

\noindent\textbf{Multi-Head Semantic Query:} Table~\ref{tab:ablation_combined}(c) shows that cluster center–guided queries significantly improve performance (78.75\% vs. 77.72\%), demonstrating enhanced semantic understanding. Queries based on initial average pooling lack the flexibility to capture rich semantic structures in natural images (78.75\% vs. 78.02\%), showing limitations compared to our cluster-guided queries. Multi-head queries outperform single-head queries (78.75\% vs. 78.41\%), confirming stronger semantic capture.

\noindent\textbf{Design Comparison:} Table~\ref{tab:ablation_combined}(d) compares CLUENet with window partitioning (from CoC) and fewer clusters ($\le$16 at all stages, from FEC). While both reduce resource consumption, they cause Top-1 accuracy drops of 0.69\% and 0.45\%, respectively, indicating that these simplifications trade off the model’s semantic modeling and representation capabilities. Additionally, increasing the number of clusters from 49 to 64 does not provide any clear performance improvement. 

\section{Conclusions} \label{sec:conclusion}
We proposed CLUENet, a novel visual model that balances strong performance with inherent interpretability. 
By introducing a Global Feature Clustering Block with efficient attention and shared assignment across the global scope, we improve performance, computational efficiency, and visualization quality.
The improved cluster pooling overcomes gradient vanishing while maintaining excellent performance and model interpretability. 
Extensive experiments on CIFAR-100 and Mini-ImageNet show that CLUENet achieves superior accuracy compared to existing clustering-based and mainstream models, while also providing clear and intuitive semantic interpretability. Future work will focus on enhancing semantic information flow across stages to improve robust recognition in complex scenarios.

\section{Acknowledgements}
This work was supported by the National Natural Science Foundation of China under Project 62572480, 62201600, 62201604, 62506371, 62522604 and 62476258.

\bibliography{refs_aaai}  

\end{document}